\documentclass[conference]{IEEEtran}
\IEEEoverridecommandlockouts

\usepackage{float}
\usepackage{tabularx,booktabs}
\newcolumntype{C}{>{\centering\arraybackslash}X} % centered version of "X" type
\usepackage{siunitx}
\usepackage{booktabs}
\usepackage{multirow}
\DeclareUnicodeCharacter{0301}{\'{e}}

\def\ps@IEEEtitlepagestyle{
  \def\@oddfoot{\mycopyrightnotice}
  \def\@evenfoot{}
}
\def\mycopyrightnotice{
  {\footnotesize xxx-x-xxxx-xxxx-x/xx/\$31.00~\copyright~2018 IEEE\hfill} % <--- Change here
  \gdef\mycopyrightnotice{}
}

\usepackage{cite}
\usepackage{amsmath,amssymb,amsfonts}
\usepackage{algorithmic}
\usepackage{graphicx}
\usepackage{textcomp}
\usepackage{xcolor}
\def\BibTeX{{\rm B\kern-.05em{\sc i\kern-.025em b}\kern-.08em
    T\kern-.1667em\lower.7ex\hbox{E}\kern-.125emX}}

\makeatletter

\def\ps@IEEEtitlepagestyle{
  \def\@oddfoot{\mycopyrightnotice}
  \def\@evenfoot{}
}
\def\mycopyrightnotice{
  {\footnotesize \hfill} % <--- Change here
  \gdef\mycopyrightnotice{}
}

\@ifundefined{showcaptionsetup}{}{
 \PassOptionsToPackage{caption=false}{subfig}}
\usepackage{subfig}
\makeatother

\usepackage{eso-pic}
\newcommand\AtPageUpperMyright[1]{\AtPageUpperLeft{
 \put(\LenToUnit{0.5\paperwidth},\LenToUnit{-1cm}){
     \parbox{0.5\textwidth}{\raggedleft\fontsize{9}{11}\selectfont #1}}
 }}
\newcommand{\conf}[1]{
\AddToShipoutPictureBG*{
\AtPageUpperMyright{#1}
}
}

\conf{} % Change according to their suggestion

\begin{document}

\title{
Word level Bangla Sign Language Dataset for Continuous BSL Recognition%{\footnotesize \textsuperscript{*}Note: Sub-titles are not captured in Xplore and
%should not be used}
}

%\iffalse
\author{\IEEEauthorblockN{Md Shamimul Islam}
%\IEEEauthorblockA{\textit{Team Leader, Perpetual Learners} \\
\textit{shamimul435@gmail.com} \\
\and
\IEEEauthorblockN{A.J.M. Akhtarujjaman Joha}
%\IEEEauthorblockA{\textit{Member, Perpetual Learners} \\
\textit{ajmjohamiu@gmail.com} 

\and
\IEEEauthorblockN{Md Nur Hossain}
%\IEEEauthorblockA{\textit{Member, Perpetual Learners} \\
\textit{nurhossainjoy69@gmail.com} 

\and
\IEEEauthorblockN{Sohaib Abdullah}
%\IEEEauthorblockA{\textit{Mentor, Perpetual Learners} \\
\textit{sohaib.abdullah2010@gmail.com} 

\and
\IEEEauthorblockN{Ibrahim Elwarfalli}
%\IEEEauthorblockA{\textit{Mentor, Perpetual Learners} \\
\textit{ieelwarfalli@mix.wvu.edu} 

\and
\IEEEauthorblockN{Md Mahedi Hasan}
%\IEEEauthorblockA{\textit{Mentor, Perpetual Learners} \\
\textit{mahedi0803@gmail.com}

}
%\fi
\maketitle

\begin{abstract}
An robust sign language recognition system can greatly alleviate communication barriers, particularly for people who struggle with verbal communication. This is crucial for human growth and progress as it enables the expression of thoughts, feelings, and ideas. However, sign recognition is a complex task that faces numerous challenges such as same gesture patterns for multiple signs, lighting, clothing, carrying conditions, and the presence of large poses, as well as illumination discrepancies across different views. Additionally, the absence of an extensive Bangla sign language video dataset makes it even more challenging to operate recognition systems, particularly when utilizing deep learning techniques.  In order to address this issue, firstly, we created a large-scale dataset called the MVBSL-W50, which comprises 50 isolated words across 13 categories. Secondly, we developed an attention-based Bi-GRU model that captures the temporal dynamics of pose information for individuals communicating through sign language. The proposed model utilizes human pose information, which has shown to be successful in analyzing sign language patterns. By focusing solely on movement information and disregarding body appearance and environmental factors, the model is simplified and can achieve a speedier performance. The accuracy of the model is reported to be 85.64\%.

\end{abstract}


\section{Introduction}
Automatic real-time Sign Language Recognition (SLR) is essential for smooth communication with deaf people due to the majority of people being unfamiliar with this language. However, real-time SLR is a challenging task as different signs may have similar hand gestures, body movements, subject viewpoints, inaccurate sign actions, etc. Many approaches have been proposed to address these
challenges over the last three decades. Kak et al.~\cite{b1} constructed the Purdue RVL-SLLL ASL database which contains a total of 1,834 signs performed by 14 signers in an indoor environment. This dataset also contains a vocabulary of 104 signs which makes 10 short stories. The RWTH BOSTON-104 and The RWTH-BOSTON-400 contain a sentence-level corpus of 50,104 and 483 signs respectively~\cite{b2}. They also introduced combined model of tangent distance and an image distortion American SLR. Furthermore, Ko et al.~\cite{b3} built a large-scale Korean continuous sign language dataset, named KETI, consists of 419 words and 105 sentences captured with 2 camera angles. They proposed translation model based on body point estimation.

\begin{figure*}[]
	\centerline
	{\includegraphics[width=180mm,scale=0.20]{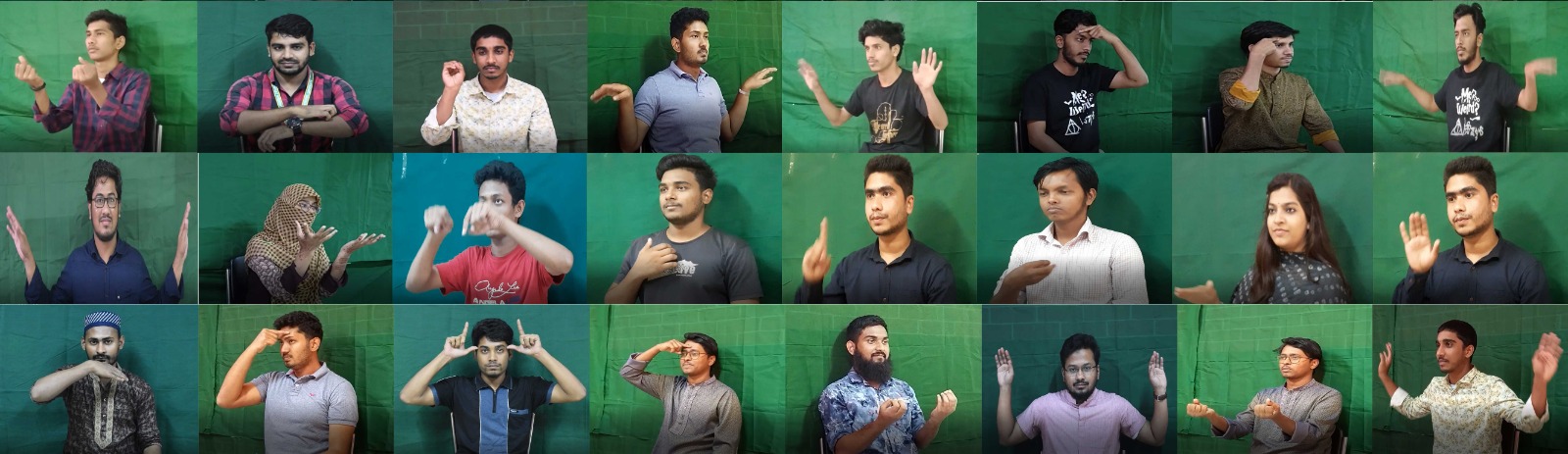}}
	\caption{Examples showcasing the variety present in our dataset, which encompasses green backgrounds, lighting situations, and signers displaying varying physical characteristics with different three angles. }
	\label{fig_1}
\end{figure*}

For Bangla Sign Language Recognition(BSLR), most of the works have been conducted on images and Bangla characters. In~\cite{bensignNet22}, a cross-correlation was applied for two-handed Bangla Sign Recognition in Bangla characters which used 80 images for 10 classes and achieved accuracy of 96\%. Rahaman et al.~\cite{bsl3612} introduced a computer vision-based contour analysis and feature-based cascaded classifier for Bangla sign language recognition which used 1,800 contour templates of 18 signs. In~\cite{bensignNet27}, authors proposed a Leap motion controller(LMC) based CNN method to to track signer's hand movements which used a limited dataset of only 14 classes. The Ishara-Lipi dataset is the first approach of multipurpose dataset of isolated Bangla characters of Bangal Sign language which consist of total 1,800 images of 36 classes~\cite{bensignNet13}. It was collected 50 images per class by different deaf and volunteer signers. Authors implemented a CNN method on Ishara-Lipi dataset and accuracy is 94.74\%. In~\cite{bensignNet11}, the 38 BdSL Dataset is alphabet-based Bangla sign image dataset which consist of 12,581 signs over 38 classes. It was collected by 310 signers collaborated with the National Federation of the Deaf people. They also built a CNN-based VGG19 architecture for recognition which achieved accuracy of 89.6\% on test data. Jim et al.~\cite{bensignNet12} proposed KU-BdSL dataset which is Bengali alphabets dataset of two variants such as, Uni-scale sign language dataset (USLD) and the Multi-scale sign language dataset (MSLD). It contains total of 1500 images of 30 classes and was collected by 33 signers with different backgrounds. Hoque et al.~\cite{b4}] proposed a dataset named BdSL36 which contains 40,000 images for 36 Bangla letters and digits and proposed a ResNet-50-based model. Miah et al.~\cite{bensignNet} proposed a BenSignNet for Bangla sign recognition, which firstly used concatenated segmentation with watershed algorithm for hand signs identification accurately and then CNN is applied for feature extraction as well as classification. They used three benchmark Bangla letter sign datasets such as ‘Ishara-Lipi’, ‘KU-BdSL’ and '38 BdSL’ and accuracy of proposed method achieved 99.06\%, 99.60\% and 94.00\%, respectively.

Due to the lack of publicly available large-scale video datasets in BSL, we introduce a large-scale BSL video dataset namely the Word level Multi-view Bangla Sign Language dataset(MVBSL-W50) with 4,000 high-resolution sign videos in basis of 50 isolated words of 13 different categories. The videos are collected with three camera angles; front left side and right side on green backgrounds which provides more opportunities to researchers to fetch new findings.  Furthermore, we also propose an attention-based sequence-to-sequence model for recognizing continuous signs in real-time scenarios. Before training, several steps have been taken to prepare data such as preparing input data, data preprocessing, normalization, and data augmentation. In preparing the input data, we obtained 1,556 landmark points per frame of each sign video using Mediapipe. Finally, the resulting sequences were then input bidirectional gated recurrent units (Bi-GRUs). We trained two models using our newly built dataset wchich obtained accuracy of 85.64\%.

\section{Word-level Multi-view Bangla Sign Language Dataset}

\begin{table}[]  \setlength\tabcolsep{34pt}
	\caption{Key statistics of MVBSL-W50 dataset}
	\begin{tabular}{ *{5}{c} }
		\hline
		\textbf{Characteristic}   & \textbf{MVBSL-W50} \\
		\hline
		Categories     &10 \\
		Words      & 50   \\
		Videos     & 4000     \\
		Avg Video Per Class      &80 \\
		Avg Video length      &2.54s\\
		Min Video length & 1.47s \\
        Max Video length & 3s  \\
        Resolution &960 x 540  \\
        Frame Rate &30fps \\
		\hline
		\label{table_1}
	\end{tabular}
\end{table}

\begin{table}[] \setlength\tabcolsep{18pt}
	\caption{Isolated word categories of MVBSL-W50 dataset}
	\begin{tabular}{ *{5}{c} }
		\hline
		%\multicolumn{2}{c}{Network} \\
		%\cline{1-3}
		\textbf{Category Name} & \textbf{Total class} & \textbf{Total videos} \\
		\hline
        Sports & 1    & 80 \\
        Food & 4     & 320 \\
		Place & 1    & 80\\
		Pronouns & 2    & 160 \\
        Animals & 3    & 240   \\
		Greetings & 4    & 320 \\
		Electronics & 1    & 80  \\
		Adjectives & 16   & 1280 \\
  		Animals & 2    & 160  \\
		Verb & 5    & 400 \\
  		Colors & 2    & 160  \\
		Days and Time  & 3    & 240 \\
        Objects of home & 8 &640 \\
        Total & 50 & 4,000 \\
		\hline
		\label{table_2}
	\end{tabular}
\end{table}

To the best of our knowledge, our newly proposed MVBSL-W50 dataset is the first large-scale video dataset in Bangla sign language which can facilitate the building of deep learning-based algorithms for BSLR. It consists of 4,000 high-resolution sign videos. In the MVBSL-W50 dataset, we carefully select words in general and emergencies basis, and choose 50 isolated words. These words belong to 13 different categories, covering the most commonly used words in BSL: adjectives, animals, Sports, colors, days and time, electronics, greetings, food, objects at home, Verb, places, pronouns. The selected words are: `bird', `bitter', `black', `book', `bread', `break', `caram', `chair', `clean', `come', `degrade', `door', `egg', `exercise', `exercise book', `fate', `february', `fish', `food', `good', `goodboy', `growth', `hearing impaired', `january', `khoda hafez', `large', `listen', `march', `me', `meat', `mobile', `more', `pencil', `picture', `procession', `quick', `remember', `rose color', `salam', `short', `sit', `small', `snake', `table', `telephone', `thanks', `tiger', `together', `up', `wet'. 

In the recordings,The signers wore clothes of different colors and performed the signs while sitting. A summary of category-wise statistics is illustrated in Table \ref{table_2}. The majority signs in the word-label are 2-4 seconds in duration, with a calculated average time of 2.30 seconds. The videos are recorded at 30 frames per second using three camera angles: front, left side, and right side, on green background which allows researchers to do background subtraction or add their own background as needed for their work. We used a Canon Power Shot A490 digital camera and smartphone cameras to capture the MVBSL-W50 dataset at an image size of $960\times540$. The tripod was placed about 2m away from the subject at a height of 1.5m. Fifty different professional sign language interpreters and volunteer signers with ages ranging from 20 to 30 years performed to record this dataset. Figure \ref{fig_1} depicts some frames of several actors in different angles. To ensure an error-free dataset, experts in Bangla sign language and professional sign language interpreters checked our video to remove possible signs of ambiguity. Table \ref{table_1} illustrates the key statistics of our newly proposed MVBSL-W50 datasets. The total number of videos is 4,000, which are split into training, development, and testing sets in the proportions of 70\%, 20\%, and 10\%, respectively which is illustrated in Table~\ref{table_3} . The average number of videos per class is 80. Table \ref{table_6} compares the different language datasets with our newly proposed MVBSL-W50 dataset. 

\begin{table*}[tb]\centering \setlength\tabcolsep{12pt}
	\caption{Comparison of different sign language datasets}
	\begin{tabular}{llllllllr}
		\hline
		\textbf{Name}   & \textbf{Lang.} & \textbf{Vocab.} & \textbf{Sentence} & \textbf{Signers} & \textbf{Multiview} & \textbf{Duration} & \textbf{Gloss} & \textbf{Pose} \\
		\hline
		Video-Based-CSL~\cite{d3} &CSL    & 178   &80 &50 &No &100 &No &Yes \\
  		Signum~\cite{d4} &DGS    & 450   &200 &25 &No &55 &Yes &No \\
		RWTH-Phoenix~\cite{d5} &DGS    & 3,000   &-- &9 &No &11 &Yes &No \\
		How 2 Sign~\cite{d6}    & ASL   & 800   &300 &327 &Yes &1 &Yes &Yes \\
		BSL Corpus~\cite{d7} &ASL    & 5,000   &-- &249 &No &-- &Yes &No \\
		KETI~\cite{b2} &CSL    & 178   &80 &50 &No &28 &No &Yes \\ 
        \hline
		MVBSL-W50(our dataset) &BSL    & 50   &- &40 &Yes &16 &No &Yes \\
		\hline
		\label{table_6}
	\end{tabular}
\end{table*}

\begin{table*}[tb]\centering \setlength\tabcolsep{14pt}
	\caption{Comparison among differnt Bangla sign language datasets}
	\begin{tabular}{llllllllr}
		\hline
		\textbf{Dataset Name}   & \textbf{Type} & \textbf{Class} & \textbf{Sample} & \textbf{Avg.} & \textbf{Multiview} & \textbf{Bg. const.} & \textbf{Level} \\
		\hline
		Rahman et al~\cite{bsl3612} &Image    & 10   &360 &36 &N &Y &Letters \\
  		Islam  et al.~\cite{d1} &Image    & 45   &30,916 &687 &N &Y &Letters \\
		Ishara-Lipi~\cite{bensignNet13} &Image    & 36   &1,800 &50 &N &Y &Letters \\
		38 BdSL~\cite{bensignNet11} &Image    & 38   &12,581 &331 &N &Y &Letters \\
		KU-BdSL~\cite{bensignNet12} &Image    & 30   &1500 &50 &N &Y &Letters \\
		BdSL36~\cite{b4} &Image    & 36   &40,000 &1111 &N &Y &Letters \\ 
        Dewanjee et al.~\cite{d2} &Video    & 10   &1,151 &12 &N &Y &Words\\
        \hline
		MVBSL-W50(our dataset)  &Video    & 50   &4,000 &80 &N &Y &Words \\
		\hline
		\label{table_7}
	\end{tabular}
\end{table*}

Table \ref{table_7} shows the comparison among Bangla sign language dataset in which most of the datasets are on image-based and letter level. However, Our dataset is first large scale word level multi-view dataset(MVBSL-W50) in Bangla sign language.  
\begin{table}[tb] \setlength\tabcolsep{11pt}
	\caption{Train, Dev and Test of MVBSL-W50 Dataset}
	\begin{tabular}{ *{5}{c} }
		\hline
		%\multicolumn{2}{c}{Network} \\
		%\cline{1-3}
		\textbf{Metric} & \textbf{Train} & \textbf{Dev} & \textbf{Test}\\
		\hline
		Number of Sign Videos & 2,800    & 800 & 400   \\
		Duration[hours] & 2.50   & .30 & .15\\
		Number of frames &     & 276000 & 138000 \\
		Number of signers & 30    & 7 & 3 \\
		Number of Camera angles & 3    & 3 & 3   \\
		\hline
		\label{table_3}
	\end{tabular}
\end{table}

\section{Methodology}
Fig.~\ref{fig_2} shows the workflow of the Bangla sign Language recognition model. To recognise the Bangla sign gestures, we proposed attention-based RNN method on our newly built dataset which are discussed in this section. We combine different stages to build model, such as, preparing input data, data preprocessing and network architecture. We start with data preparing of video of our dataset to find out the suitable sign language representative body joint information. Then, we preprocess the pose information to convert into feature vector with timestep. After that, we augment the sequence by overlapping the clips. Finally, we discuss the attention-based RNN networks and evaluate the results.

\subsection{Preparing Input Data}
In this work, we have extracted the body landmarks by applying a multistage pipeline called MediaPipe Holistic [8], which directly proposes 1556 points from hands, face, and pose. We collected all the points of hand and face landmarks, whereas we have picked selective points of pose because we searched for those points that contain rich sign representation capacity. The selective points of pose are right thumb, left thumb, right index, left index, right pinky, left pinky, right wrist, left wrist, right elbow, left elbow, right shoulder, left shoulder, and nose.
In this experiment, we found that landmarks below the human hip do not have a notable impact on sign language. Usually, hip points do not show any major changes when a sign occurs. Similarly, both legs do not need any movement during sign language activities. On the other hand, hand and face points show a strong pattern during sign language. Finally, we found that face, hand, and selective points of pose can construct more robust and representative sign language descriptors than others. Therefore, we selected 1556 points as our effective sign language descriptors.

\begin{equation}
  \boldsymbol{{x}}  = (x_{1} , x_{2} , x_{3} , ..... , x_{1556} )
  \label{eq_1}
\end{equation}

\begin{equation}
     \begin{bmatrix}
     x^{1}_{1} & x^{1}_{2} & x^{1}_{3}& \cdots & x^{1}_{1556}\\
     x^{2}_{1} & x^{2}_{2} & x^{2}_{3}& \cdots & x^{2}_{1556}\\
     x^{3}_{1} & x^{3}_{2} & x^{3}_{3}& \cdots & x^{3}_{1556}\\
     \vdots & \vdots & \vdots & \vdots & \vdots \\
     x^{30}_{1} & x^{30}_{2} & x^{30}_{3}& \cdots & x^{30}_{1556}\\     
     \end{bmatrix}=
     \begin{bmatrix}
     \boldsymbol{x_{1}} \\
     \boldsymbol{x_{2}} \\
     \boldsymbol{x_{3}} \\
     \vdots  \\
     \boldsymbol{x_{30}} \\     
     \end{bmatrix} = \boldsymbol{X} \epsilon \mathbb{R}^{30\times1556} 
      \label{eq_2}
\end{equation}

\begin{equation}
  \boldsymbol{V} = ({\boldsymbol{X}_{1}} , {\boldsymbol{X}_{2}} , {\boldsymbol{X}_{3}} ,....., \boldsymbol{X}_{T} )
   \label{eq_3}
\end{equation}

\subsection{Data Preprocessing}
We have obtained 1556 body points per frame from MediaPipe Holistic. Points that were not detected by MediaPipe were given a value of 0. By selecting points from hands, face, and pose, we created a 1556-dimensional vector of landmark points from a single frame. We then divided an entire sign language video into timesteps of 30 frames each and used equations \ref{eq_1}, \ref{eq_2}, and \ref{eq_3} to explain how we segmented the video using sequences of landmark points. In this case, $ x_{i} $ represents the 1556-dimensional landmark points vector for each frame, \textbf{X} represents the feature vector for each timestep, \textbf{V} represents the sequence of features for the sign language video, and T represents the total number of timestep sequences in the video.

\subsection{Network Architecture}
\begin{figure}[]
	{\includegraphics[width=\columnwidth,height=.4\textheight]{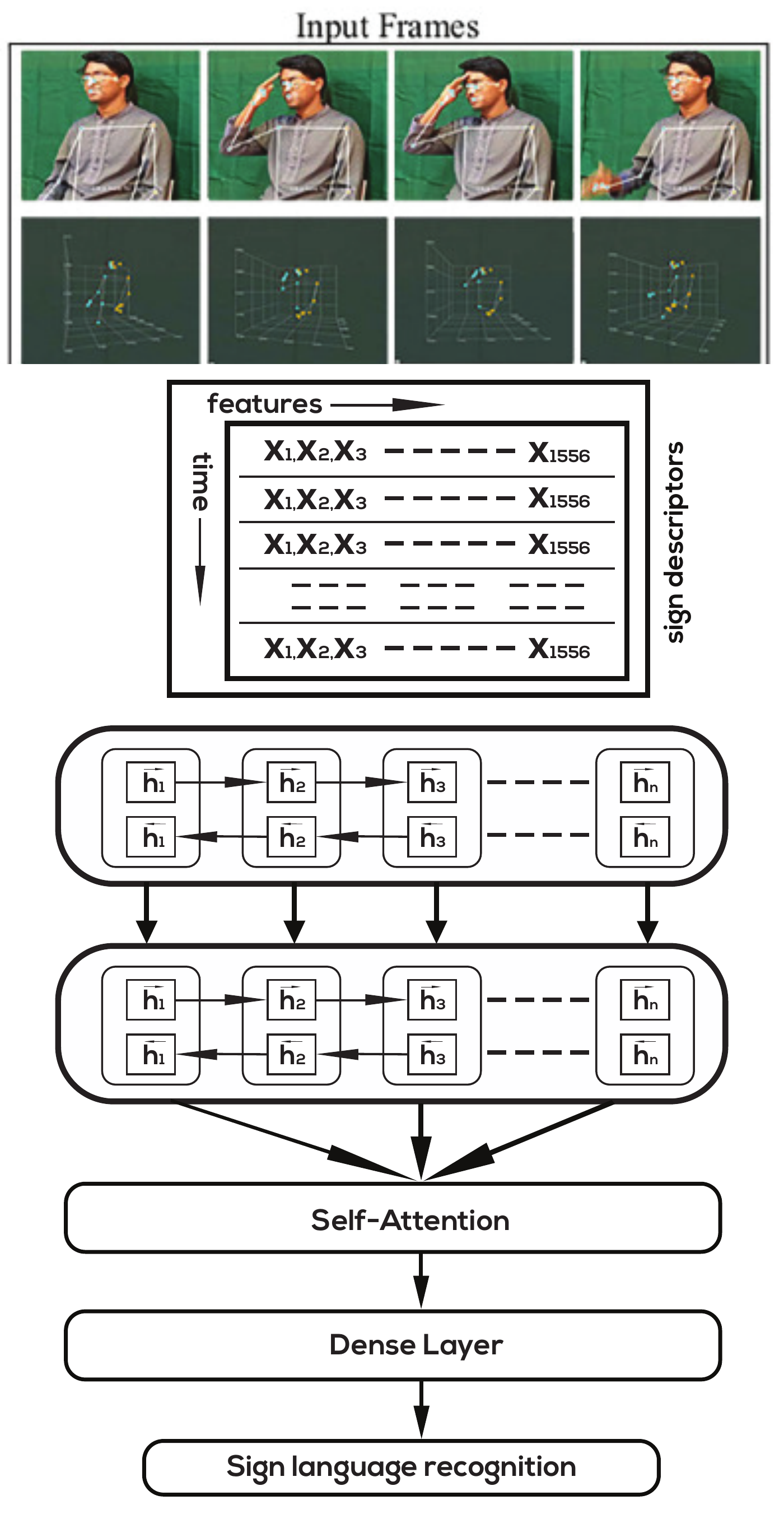}}
	\caption{The overview of our proposed method for BSLR.}
	\label{fig_2}
\end{figure}

In our experiment, we examined different attention-based and non-attention-based RNN architectures, such as bidirectional long short-term memory (Bi-LSTM), bidirectional gated recurrent units (Bi-GRU), long short-term memory (LSTM)~\cite{lstm}, and gated recurrent units (GRU). From this, we found that the attention-based Bi-GRU is an effective network for BSL, as depicted in Figure~\ref{fig_2}. Table~\ref{table_4} shows the results of the different architectures of RNN-based models. The GRU and LSTM enable the adaptively intercepting of dependencies from large data sequences without disregarding information from previous parts of the sequence. The bidirectional merge the forward and backward hidden layers, enabling them to perform each sequence in both left-to-right and right-to-left directions, and embed the sequential dependencies in both directions. At first, we used two different recurrent units, each with 128 and 256 units, that take \textbf{V} as input and then make an intermediate representation. After that, this representation is sent as a feed to the attention layer for both networks. The attention layer allows the adaptively highlighting of the most important parts of the sequence, allowing it to better capture the relevant information and improve its performance on the task. This can lead to more accurate and relevant predictions, especially for long and complex data sequences. Finally, we used two dense layers in experiment networks.   

% \begin{table}
% 	\caption{Training summary of our attention-based RNN network. \label{table:training_details}}
% 	\begin{tabular*}{21pc}{@{\extracolsep{\fill}}ll@{}}
% 		\hline \noalign{\vspace{3pt}}
% 		\textbf{Hyper-parameter} &\qquad \textbf{Value} \\ [3pt] \hline\noalign{\vspace{3pt}}
% 		Optimizer     			&\qquad RMSprop~\cite{b5} \\[3pt]
% 		Objective function  	&\qquad Categorical Cross-entropy\\[3pt]
% 		%Epochs        			&\qquad $ 200 $ \\ [3pt]
% 		Initial learning rate	&\qquad $10^{-4}$  \\[3pt]
% 		Mini-batch size			&\qquad $ 512 $ \\
% 		\hline
% 	\end{tabular*}
% \end{table}

\begin{table*}[h!]\centering
	\caption{Weighted average precision, weighted average recall, weighted average F-1, accuracy  \& Cohen's Kappa score values are compared for different RNN architectures for continous Bangla sign language. \label{table:training_details}}
	\begin{tabular}{llllllllr}
		\hline
		%\multicolumn{2}{c}{Network} \\
		%\cline{1-3}
		{RNN Network Name} & {RNN Hidden Units} & {Attention Units} & {Epocs} & {Precision} & {Recall} & {F-1 Score} &{Accuracy} & {Cohen's Kappa}  \\
		\hline
		LSTM & 128, 256 &128 & 300    & 83.52  & 82.93 & 82.98 & 82.93  & 82.56  \\ %model___07
		GRU & 128, 256 &128 & 300    & 84.30  & 83.57 & 83.63 & 83.57  & 83.22  \\ %model 1-5
		BiLSTM & 128, 256 &128 & 300    & 84.84  & 84.53 & 84.53 & 84.53  & 84.19  \\ %model_05
		BiGRU & 128, 256 &128 & 300    &85.91  & 85.27 & 85.29 & 85.27  & 84.95  \\
		LSTM + Attention & 128, 256 &128 & 300    & 84.40  & 83.64 & 83.66 & 83.64  & 83.28  \\  %model 1-5
		GRU + Attention & 128, 256 &128 & 300    & 83.49  & 82.93 & 82.89 & 82.93  & 82.56  \\ %model___08
		BiLSTM + Attention & 128, 256 &128 & 300    & 84.84  & 84.53 & 84.53 & 84.53  & 84.19 \\ %model_05
		BiGRU + Attention & 128, 256 &128 & 300    & \textbf{85.97}  & \textbf{85.64} & \textbf{85.64} & \textbf{85.64}  & \textbf{85.33}  \\
		\hline
		\label{table_4}
	\end{tabular}
\end{table*}
%128, 256 &128 & 300    & 84.30  & 83.57 & 83.63 & 83.57  & 83.22  \\ %model 1-5

\begin{figure}[]
	{\includegraphics[width=\columnwidth,height=.15\textheight]{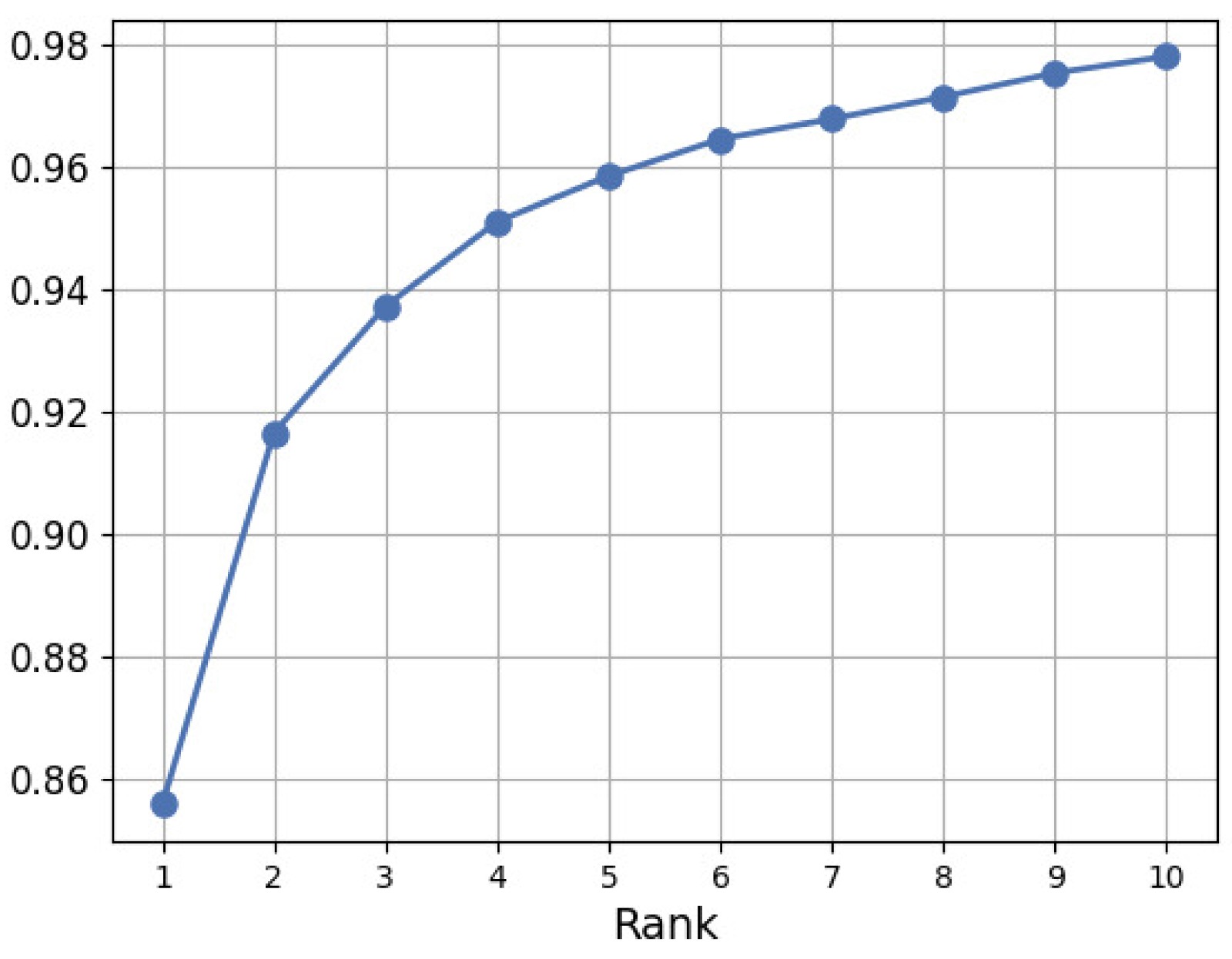}}
	\caption{The rank-1 to rank-10 sign language accuracy using the attention-based BiGRU Model.}
	\label{fig_3}
\end{figure}

\begin{figure}[]
	{\includegraphics[width=90mm,scale=0.30]{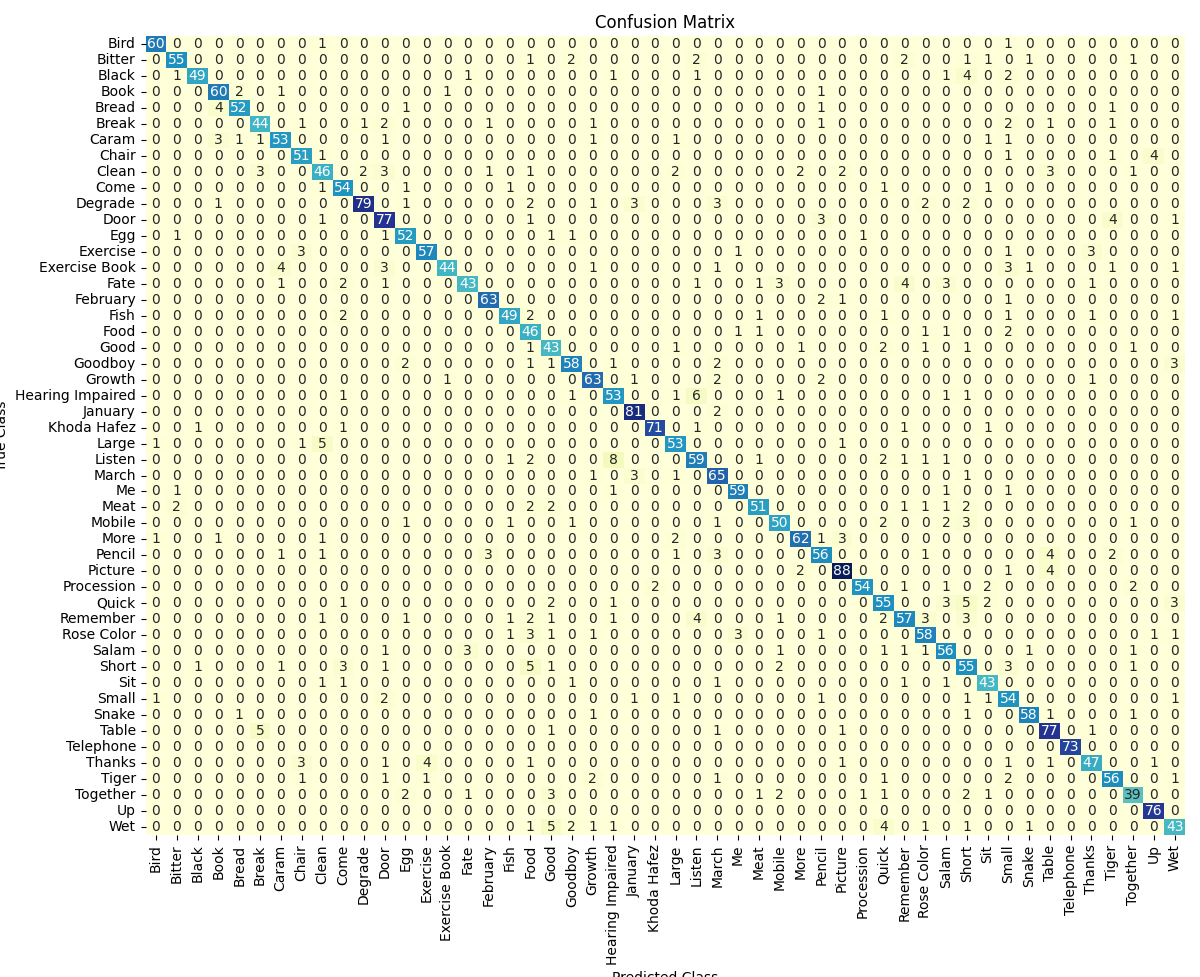}}
	\caption{Confusion matrix showing Bangla sign language recognition results with 50 classes. Rows show the actual class of a repetition and columns show the classifier's prediction. }
	\label{fig_4}
\end{figure}

\section{EXPERIMENTAL RESULTS}
\subsection{Dataset}
In this study, we used two newly built datasets, namely MVBSL-W50 which are described in section II.

\subsection{Implementation Details}
We utilized the Keras deep learning package provided by TensorFlow~\cite{tensorflow} to create our attention-based and without attention-based RNN architectures. The NVIDIA GeForce GTX 1050 GPU was used to conduct our experiment. We divided the datasets into 80\% for training and 20\% for testing. Our proposed model was trained using the categorical cross-entropy loss function and the RMSprop optimization algorithm~\cite{rmsprop}. To initialize the weights in our model, we employed the Xavier initialization. We trained our model over 200 epochs. 

\subsection{Results}
attention-based and without attention-based various recurrent neural networks(RNNs) were applied in our experiment and the results of them are described in the table ~\ref{table_4}. Bi-GRU network and attention-based Bi-GRU network demonstrated results on our dataset. Bi-GRU model achieved 85.27\% accuracy and 85.91\% precision on test data, while the  attention-based Bi-GRU achieved 85.64\% accuracy and 85.97\% precision on test data. Our effective attention-based BiGRU model achieved a rank-1 accuracy of 85.64\% and rank-10 accuracy of about 98.00\% in Figure~\ref{fig_3}.

Some words were incorrectly recognized by our best-resulting BiGRU model, such as `black', `clean', `exercise book', `fate', `february', `large', `short', `table', `meat', `thank' and `wet'. Upon closer observation, we found some issues that caused these recognition errors, primarily that most of these signs are formed using two hands fisting the fingers. As a result, we failed to capture the finger point landmarks in Google MediaPipe, which contributed to null points in our \textbf{V} (the sequence of features for a sign video). Additionally, some Bangla sign movements are similar to those of other words, making them more challenging to recognize. Despite these challenges, our BiGRU model outperformed in most words, achieving 100\% accuracy as shown in Figure~\ref{fig_4}.
% \begin{table}
% 	\caption{Result summary in experiments. \label{table:training_details}}
% 	\begin{tabular*}{21pc}{@{\extracolsep{\fill}}lll@{}}
% 		\hline 
% 		\textbf{Network Name} & \textbf{Training Accuracy} & \textbf{Test Accuracy} \\
%        \hline
% 		Bi-GRU     & 90\% & 88\% \\
% 		Bi-LSTM  	& 87\% & 85\%\\
% 		\hline
% 	\end{tabular*}
% \end{table}

\section{Conclusion}
In this work, we have introduced a multi-view large scale  word level BSL dataset (MVBSL-W50) as well as a BSL continuous sign language recognition model which truly demonstrates the capabilities for it which truly demonstrates the capabilities that would be deployed in further research. We considered human pose information as a highly effective sign descriptor because of its ability to not only provide strong sign representation but also shows robustness to several challenges of sign recognition such as changes in viewing angle, lighting, clothing, and carrying conditions. In the future, we plan to utilize a more precise pose estimation algorithm to enhance our performance when dealing with significant changes in viewing angles. Additionally, we intend to incorporate a more advanced and accurate sign recognition network that will be applicable to practical purposes, such as real-time sign translation.

%Ctrl + t 
%\vspace{12pt}
%\color{red}
%IEEE conference templates contain guidance text for composing and formatting conference papers. Please ensure that all template text is removed from your conference paper prior to submission to the conference. Failure to remove the template text from your paper may result in your paper not being published.
%Crtl + u
\end{document}